\documentclass[conference]{IEEEtran}
\pdfoutput=1
\IEEEoverridecommandlockouts
\usepackage{float}
\usepackage{caption}
\usepackage{times}

\usepackage[numbers]{natbib}
\usepackage{multicol}
\usepackage[bookmarks=true]{hyperref}
\usepackage{multirow}
\usepackage{booktabs}
\usepackage{makecell}
\usepackage{fleqn}
\usepackage{tabularx}
\usepackage{siunitx}
\usepackage{amsmath}
\usepackage{stfloats}
\usepackage{cuted}
\usepackage{placeins}
\usepackage{graphicx}
\usepackage{xcolor}

\newcommand{\ie}{\textit{i}.\textit{e}. }
\newcommand{\eg}{\textit{e}.\textit{g}. }

\newcommand{\snuggg}{\vspace{-3em}}
\newcommand{\snugg}{\vspace{-2em}}

\begin{document}

\title{Never Stop Learning: The Effectiveness of Fine-Tuning in Robotic Reinforcement Learning}

\author{
    \authorblockN{
        Ryan Julian\authorrefmark{2}\authorrefmark{3},
        Benjamin Swanson\authorrefmark{2},
        Gaurav S. Sukhatme\authorrefmark{3},
        Sergey Levine\authorrefmark{2}\authorrefmark{4},
        Chelsea Finn\authorrefmark{2}\authorrefmark{5} and
        Karol Hausman\authorrefmark{2}
    }
    \authorblockA{\authorrefmark{2}Google Research, Robotics at Google Team}
    \authorblockA{\authorrefmark{3}Department of Computer Science, University of Southern California}
    \authorblockA{\authorrefmark{4}Department of Electrical Engineering and Computer Sciences, University of California, Berkeley}
    \authorblockA{\authorrefmark{5}Department of Computer Science, Stanford University}
}

\maketitle
\noindent
\snuggg
\begin{strip}
  \begin{minipage}{\textwidth}
  \vspace{-1cm}
    \begin{figure}[H]
        \centering
        \noindent
        \hspace{-0.2cm}\includegraphics[width=\textwidth]{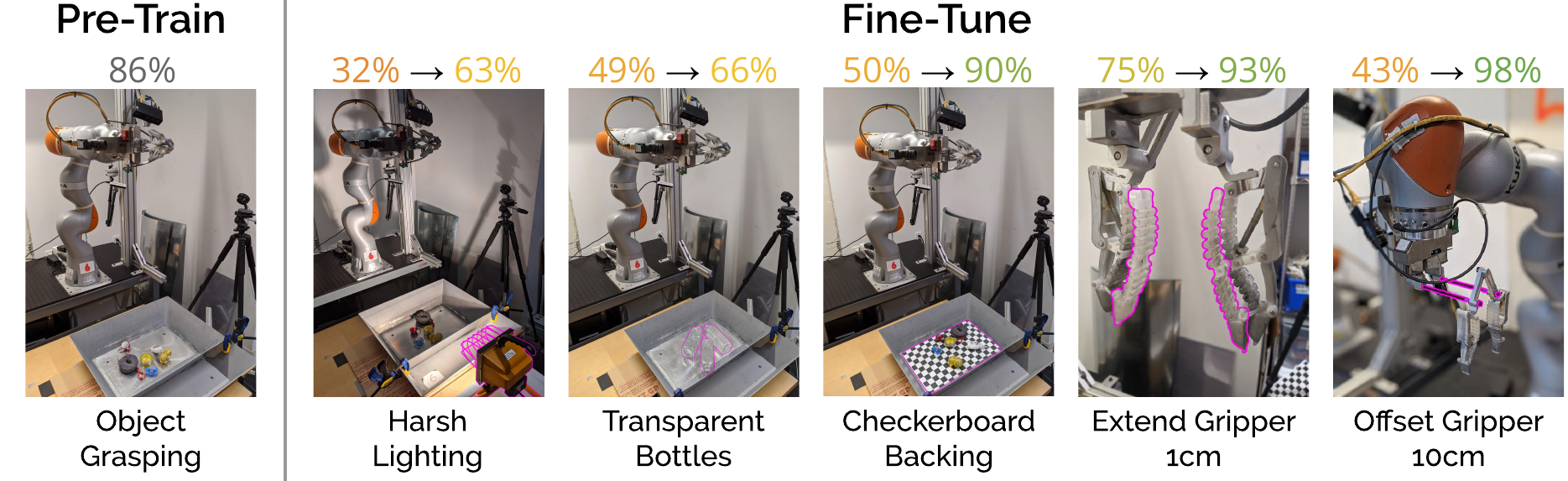}
          \caption{
          Original robot configuration used for pre-training (left), and adaptation challenges (highlighted in pink) studied in this work (right) with associated performance improvements (top) obtained using our fine-tuning method.
          \label{fig:headline}}
    \end{figure}
  \end{minipage}
\end{strip}
\snugg

\begin{abstract}
One of the great promises of robot learning systems is that they will be able to learn from their mistakes and continuously adapt to  ever-changing environments.
Despite this potential, most of the robot learning systems today are deployed as a fixed policy and they are not being adapted after their deployment.
Can we efficiently adapt previously learned behaviors to new environments, objects and percepts in the real world?
In this paper, we present a method and empirical evidence towards a  robot learning framework that facilitates continuous adaption.
In particular, we demonstrate how to adapt vision-based robotic manipulation policies to new variations by fine-tuning via off-policy reinforcement learning, including changes in background, object shape and appearance,
lighting conditions, and robot morphology. Further, this adaptation uses less than 0.2\% of the data necessary to learn the task from scratch.
We find that the simple approach of fine-tuning pre-trained policies leads to substantial performance gains over the course of fine-tuning, and that pre-training via RL is essential: training from scratch or adapting from supervised ImageNet features are both unsuccessful with such small amounts of data. We also find that these positive results hold in a limited continual learning setting, in which we repeatedly fine-tune a single lineage of policies using data from a succession of new tasks.
Our empirical conclusions are consistently supported by experiments on simulated manipulation tasks, and by 52 unique fine-tuning experiments on a real robotic grasping system pre-trained on 580,000 grasps.
For video results and an overview of the methods and experiments in this study, see the project website at {\color{blue}\url{https://ryanjulian.me/continual-fine-tuning}}.

\end{abstract}

\maketitle

\section{Introduction}
\label{sec:introduction}
The ability to constantly learn, adapt, and evolve is arguably one of the most important properties of an intelligent agent prepared to exist in the real world.  
Similarly, our robots should be able to continuously learn and adapt throughout their lifetime to the ever-changing environments that they are deployed in.
This is a widely recognized requirement. 
In fact, there is an entire academic sub-field of lifelong learning~\cite{thrun1998} that is interested in the problem of agents that never stop learning.
Despite the wide interest in this ability, most of the intelligent agents deployed today are not tested for their adaptation capabilities.
Even though techniques such as reinforcement learning theoretically provide the ability to perpetually learn from trial and error, this is not how they are typically evaluated.
Instead, the predominant method of acquiring a new task with reinforcement learning is to initialize a policy from scratch, collect entirely new data in a stationary environment, and evaluate a static policy that was trained with this data.

This static paradigm does not evaluate the robot's capability to adapt.
It also traps robotic reinforcement learning in the worst-case regime for sample efficiency: the cost to acquire a new task is dominated by sample efficiency of the learning algorithm and the complexity of the task, as reflected in cost of acquiring diverse task data starting from na\"{i}ve (e.g. random) exploration.

Most machine learning models successfully deployed in the real world, such as those used for computer vision and natural language processing (NLP) do not live in this regime.
For instance, the predominant method of acquiring a new computer vision task is to start learning the new task with a pre-trained model for a related task, acquired from a pre-collected data set, and \textit{fine-tune} that model to achieve the new task~\cite{donahue2014decaf,ulmfit,devlin2018bert}.
This changes the sample efficiency regime of the learning process from one which is dominated by \textit{task complexity} to one that is dominated by \textit{task novelty}, i.e. the difference between the new task and the task on which the model was pre-trained. 
While a number of works have studied how to use pre-trained ImageNet~\cite{deng2009imagenet} features for robotics~\cite{yosinski2014transferable,huh2016makes,kornblith2019better},
there are remarkably few works that study how to adapt motor skills
themselves. Our work attempts to bridge this gap.

We adapt an image-based grasping policy to changes in background, object shape and appearance, 
lighting conditions, and robot morphology and kinematics, while using less than 0.2\% of the data necessary to learn the same task from scratch (see Fig.~\ref{fig:headline}).
Our results, supported by simulation and extensive real-world experiments, indicate that a pre-adaptation policy acquired for a task using reinforcement learning can be used to acquire policies for nearby tasks using very little new data and a simple update procedure.
Furthermore, we find that this approach of adapting pre-trained policies  with off-policy reinforcement learning (RL) leads to substantial improvements over the course of fine-tuning, and that pre-training via RL is essential: it significantly outperforms conventional pre-training techniques using supervised learning on task-agnostic datasets.
We believe this simple adaptation scheme provides a promising solution for creating a lifelong learning robotic agent, and show this potential using a simple continual learning experiment.

\textbf{The main contributions of this work are (1) a careful real-world study of the problem of end-to-end skill adaptation for a continuously-learning robot, and (2) evidence that a very simple fine-tuning method can achieve that adaptation.}

Instead of focusing on the robot's performance in the environment in which it was trained, we purposefully modify the robot and its environment, characteristic of the persistent change of the real world, and investigate its ability to adapt.
Likewise, rather than proposing a new adaptation algorithm, with new complexity and caveats, we show how to successfully adapt robotic policies to substantial changes, using only the most basic components of existing off-policy reinforcement learning algorithms. 
\textbf{To our knowledge, this work is the first to demonstrate that simple fine-tuning of off-policy reinforcement learning can successfully adapt to substantial task, robot, and environment variations which were not present in the original training distribution (\ie off-distribution).}

\section{Related Work}
\label{sec:related_work}
Reinforcement learning is a long-standing approach for enabling robots to autonomously acquire skills~\cite{kober2013reinforcement} such as locomotion~\cite{peter_stone2004,tedrake_applied_2004}, pushing objects~\cite{mahadevan1992automatic,finn2017deep}, ball-in-cup manipulation~\cite{kober2009policy}, peg insertion~\cite{gullapalli1994acquiring, levine2016end, schoettler2019deep, lee2018making, zeng2018learning}, throwing objects~\cite{ghadirzadeh2017deep, zeng2019tossingbot}, and grasping~\cite{pinto2016supersizing,kalashnikov2018scalable}. We particularly focus on the problem of deep reinforcement learning from raw pixel observations~\cite{levine2016end}, as it allows us to place little restrictions on state representation. A number of works have also considered this problem setting~\cite{finn2016deep,ghadirzadeh2017deep,finn2017deep,zeng2018learning,agrawal2016learning,mnih2015human}. However, a key challenge with deep RL methods is that they typically learn each skill from scratch, disregarding previously-learned skills. If we hope for robots to generalize to a broad range of real world environments, this approach is not practical.

We instead consider how we might transfer knowledge for efficient learning in new conditions~\cite{taylor2009transfer, pan2009survey, Tan_2018}, a widely-studied problem particularly outside of the robotics domain~\cite{donahue2014decaf,ulmfit,devlin2018bert,dai2007boosting,raina2007self}. Prior works in robotics have considered how we might transfer information from models trained with supervised learning on ImageNet~\cite{deng2009imagenet} by fine-tuning~\cite{levine2016end,finn2016deep,gupta2018robot,pinto2016supersizing} or other means~\cite{Sermanet2017Rewards,hazara2019transferring}. Our experiments show that transfer from pre-trained conditions is significantly more successful than transfer from ImageNet. Other works have leveraged experience in simulation~\cite{Sadeghi_2017,Tobin_2017,SadeghiTJL18,sim2real,openai2019solving,Rusu2016SimtoRealRL,peng2018sim,higuera2017adapting,hamalainen2019affordance} or representations learned with auxiliary losses~\cite{Riedmiller2018LearningBP,mirowski2016learning,sax2019mid} for effective transfer. While successful, these approaches either require significant engineering effort to construct an appropriate simulation or significant supervision. Most relevantly, recent work in model-based RL has used predictive models for fast transfer to new experimental set-ups~\cite{chatzilygeroudis2018using,ha2018recurrent}, \ie by fine-tuning predictive models~\cite{dasari2019robonet}, via online search of a pre-learned representation of the space models, policies, or high-level skills~\cite{chatzilygeroudis2018reset,cully2015robots,kaushik2020adaptive,merel2019reusable}, or by learning physics simulation parameters from real data~\cite{rastogi2018sample, jeong2019modelling}. We show how fine-tuning is successful with a model-free RL approach, and show how a state-of-the-art grasping system can be adapted to new conditions.

Other works have aimed to share and transfer knowledge across tasks and conditions by simultaneously learning across multiple goals and tasks~\cite{ruder2017overview}. For example, prior works in model-based RL~\cite{finn2017deep,yen2019experience,nagabandi2019deep} and in goal-conditioned RL~\cite{agrawal2016learning,nair2018visual,Pathak_2018,pong2019skew,yu2019unsupervised} have shared data and representations across multiple goals and objects. Along a similar vein, prior work in robotic meta-learning has aimed to learn representations that can be quickly adapted to new dynamics~\cite{clavera2018learning,alet2018modular,nagabandi2018deep} and objects~\cite{finn2017one,james2018task,yu2018one,bonardi2019learning}. We consider adaptation to a broad class of changes including dynamics, object classes, and visual observations, including conditions that shift substantially from the training conditions, and do not require the full set of conditions to be represented during the initial training phase.

\section{The Robustness of Learned Policies: A Case Study}
\label{sec:case_study}
To study the problem of adaptation, we utilize a grasping policy pre-trained with RL, which we evaluate in five different conditions that were not encountered during pre-training. 
In this section, we will describe the pre-training process and test the robustness of the pre-trained policy to various robot and environment modifications. We choose these modifications to reflect changes we believe a learning robot would experience, and should be expected to a adapt to, when deployed ``on the job'' in the real world. In Section~\ref{sec:experiments}, we will describe a simple fine-tuning based adaptation process, and evaluate it using these modifications.

\subsection{Pre-training process}
\label{sec:pretraining}
We pre-train the grasping policy, which we refer to as the ``base policy,'' using the QT-Opt algorithm in two stages, as described in~\cite{kalashnikov2018scalable}.
First, we train a Q-function network offline using data from 580,000 real grasp attempts over a corpus of 1,000 visually and physically diverse objects.
Second, we continue training this network online\footnote{Following the example set by~\cite{kalashnikov2018scalable}, we refer this procedure as ``online'' rather than ``on-policy,'' because the policy is still updated by the off-policy reinforcement learning algorithm} over the course of 28,000 real grasp attempts on the same corpus of objects.
That is, we use a real robot to collect trials using the current network, update the network using these new trials, deploy the updated network to the real robot, and repeat.
This procedure yields a final base policy that achieves 96\% accuracy on a set of previously-unseen test objects.
We use a challenging subset of six of these test objects for most experiments in this work. On this set, \textbf{our base model achieves a success rate of 86\% on the baseline grasping task}.

\subsection{Robustness of the pre-trained policy}
We begin by choosing set of significant modifications to the robot and environment, which we believe are characteristic of a real-world continual learning scenario. We then evaluate the performance of the base policy on increasingly-severe versions of these modifications. This process allows us to assess the limits of robustness of policies trained using the pre-training method. Once we find a modification that is sufficiently-severe to compromise the base policy's performance in each category, we use it to define a ``Challenge Task'' for our study of adaptation methods.

\begin{table}[t]
    \centering
    \setlength{\tabcolsep}{1em}
    \begin{tabularx}{\linewidth}{llll*{9}{c}}
        \toprule
        \multicolumn{1}{l}{Challenge Task} & \multicolumn{1}{l}{Type} & \multicolumn{1}{l}{Base Policy} & \multicolumn{1}{l}{$\Delta$} \\
        \midrule
        Checkerboard Backing        & Background           & 50\%      & -36\%  \\
        Harsh Lighting              & Lighting conditions  & 31\%      & -55\% \\
        Extend Gripper \SI{1}{cm}   & Gripper shape        & 76\%      & -10\% \\
        Offset Gripper \SI{10}{cm}  & Robot morphology     & 47\%      & -39\%  \\
        Transparent Bottles         & Unseen objects       & 49\%      & -37\% \\
        \bottomrule
    \end{tabularx}
    \caption{
    Summary of modifications to the robot and environment, and their effect on the performance of the base policy. Changing the background lighting, morphology, and objects leads to substantial degradation in performance compared to the original training conditions.
    }
    \label{tab:modifications}
\end{table}

\begin{figure}
  \noindent
  \centering
  \includegraphics[width=\columnwidth]{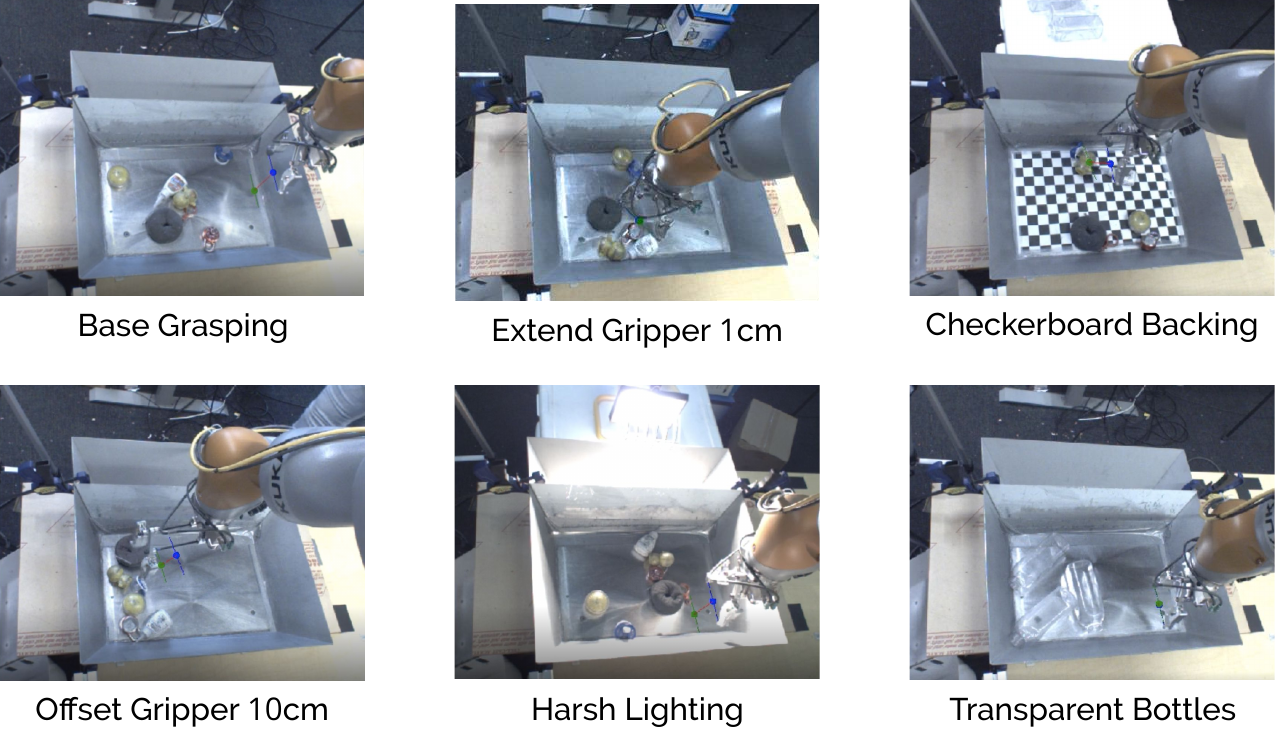}
  \caption{
  Views of from the robot camera for each of our six Challenge Tasks and the base grasping task.
  }
  \label{fig:kcam}
\end{figure}

Next, we describe these challenges and the corresponding performance of the base policy.

\textbf{Background:}
We introduce a black-white \SI{1}{inch} checkerboard pattern that we glue to the bottom of the robot's workspace (see Fig.~\ref{fig:headline}, fourth from left). We observe that conventional variations in the workspace surface, such as uniform changes in color or specularity, have no effect on the base policy's performance. Introducing an checkerboard pattern often fools the robot into grasping at checkerboard edges rather than objects. This adversarial modification compromises the base policy's performance to 50\% (-36\% compared to the base task).

\textbf{Lighting conditions:}
We introduce a high-intensity halogen light source parallel with the workspace (see Fig.~\ref{fig:headline}, second from left), creating a bright spot in the robot's camera view, and intense light-dark contrasts along the plane of the workspace. The base policy was trained in standard indoor lighting conditions, with no exposure to natural light or significant variation. We observe that mild perturbations in lighting conditions (i.e. those which can be created by standard-intensity household lights) have no effect on the base policy's performance. Using the very bright halogen light source has a severe impact, and degrades the base policy's performance to 31\% (-55\% compared to the baseline).

\textbf{Gripper shape:}
We extend the parallel gripper attached to the robot by \SI{1}{\cm} and significantly narrow its width and compliance in the process (see Fig.~\ref{fig:headline}, fifth from left). This changes the robot's kinematics (lengthening the gripper in the distal direction), while also lowering the relative pose of the robot with respect to the workspace surface by \SI{1}{\cm}. This modification compromises the base policy's performance to 76\% (-10\% compared to the baseline).

\textbf{Robot morphology:}
We translate the gripper laterally by \SI{10}{\cm} (see Fig.~\ref{fig:headline}, far-right). Note that during training this policy experienced absolutely no variation in robot morphology. We observe that translating the gripper laterally by up to \SI{5}{\cm} has no impact on performance. By translating the gripper laterally by \SI{10}{\cm} (approximately a full gripper or arm link width), we degrade the base policy's performance to 47\% (-39\% compared to the baseline).

\textbf{Unseen objects:}
We introduce completely-transparent plastic beverage bottles (see Fig.~\ref{fig:headline}, third from left) that were not present in the training set. Based on our experiments, the system is robust to a broad variety of previously-unseen objects, as long as they have significant opaque components. For example, even though there are no drinking bottles in the training set, we find the system is able to pick up labeled drink bottles with 98\% success rate. Success rates for other novel, opaque objects are similarly consistent with the baseline performance on the test set. However, we find that introducing completely-transparent drink bottles causes the base policy to often grasp where two bottles are adjacent, \ie as though it cannot differentiate which parts of the scene are inside vs outside a bottle. By introducing completely-transparent plastic beverage bottles, we are able to compromise the base policy's performance to 49\% (-37\% compared to the baseline).

See Table~\ref{tab:modifications} for a summary of the modification experiments, and their effect on base policy performance.

\section{Large-Scale Experimental Evaluation}
\label{sec:experiments}
\begin{figure}
  \noindent
  \centering
  \includegraphics[width=\columnwidth]{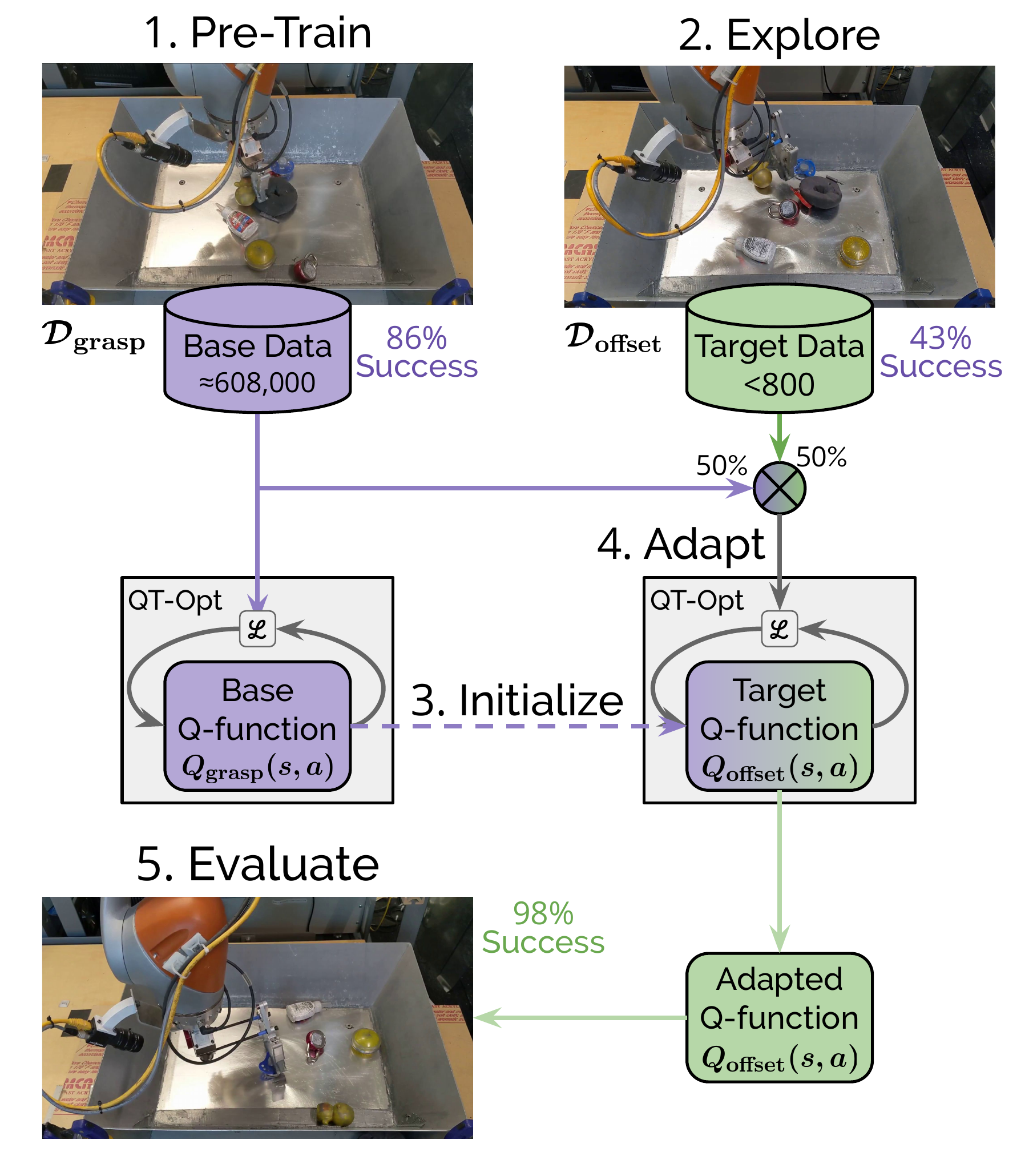}
  \caption{
  Schematic of the simple method we test in Section~\ref{sec:experiments}, using the conceptual framework we discuss in Appendix~\ref{sec:fine_tuning_framework}. We pre-train a policy using the old data from the pre-training task, which is then adapted using the new data from the fine-tuning task.
  }
  \label{fig:flowchart}
\end{figure}
We define then evaluate a simple technique for offline fine-tuning.

Our experiments model an ``on the job'' adaptation scenario, where a robot is initially trained to perform a general task (in our case, grasping diverse objects), and then the conditions of the task change in a drastic and substantial way as the robot performs the task, \eg through the introduction of significantly brighter lighting, or a peculiar and unexpected type of object.
The robot must adapt to this change quickly in order to recover a proficient policy. 
Handling these changes reflects what we expect to be a common requirement of reinforcement learning policies deployed in the real world: since an RL policy can learn from all of the experience that it has collected, there is no need to separate learning into clearly distinct training and deployment phases.
Instead, it is likely desirable to allow the policy to simply continue learning ``on the job'' so as to adapt to these changes.

\subsection{A very simple fine-tuning method}
We define a very simple fine-tuning procedure for off-policy RL, as follows (Fig.~\ref{fig:flowchart}).

First, we (1) pre-train a general grasping policy, as describe in Section~\ref{sec:pretraining} and~\cite{kalashnikov2018scalable}. To fine-tune a policy onto a new target task, we (2) use the pre-trained policy to collect an exploration dataset of attempts on the target task; then (3) initialize the same off-policy reinforcement learning algorithm which was used for pre-training (QT-Opt, in our case) with the parameters of the pre-trained policy, and both the target task and base task datasets\footnote{We assume this dataset was saved during training of the base policy} as the data sources (\eg replay buffers); we then (4) update the policy with this training algorithm, using a reduced learning rate, and sampling training examples with equal probability from the base and target task datasets, for some number of update steps. Finally, we (5) evaluate the fine-tuned policy on the target task.

Our method is offline, \ie it uses a single dataset of target task attempts, and requires no robot interaction after initial dataset collection to compute a fine-tuned policy, which may then be deployed onto a robot.

\subsection{Evaluating offline fine-tuning for real-world grasping}
We now turn our attention to how to evaluating this simple method's effectiveness as an adaptation procedure for end-to-end robot learning, and perhaps continual learning. Our goal is to determine whether the method is sample efficient, whether it works over a broad range of possible variations, and to determine whether it performs better than simpler ways of acquiring the target tasks.

With this goal in mind, we conduct a large panel of ablation experiments experiments on a real 7 DoF Kuka arm. These experiments evaluate the performance of our method across the diverse range of previously-defined Challenge Tasks and a continuum of target task dataset sizes, and compare this performance to two comparison methods.

The experiments are very challenging. The Transparent Bottles task in particular presents a major challenge to most grasping systems: the transparent bottles generally confuse depth-based sensors and, especially in cluttered bins, require the robot to singulate individual items and position the gripper in the right orientation for grasping. Although our base policy uses only RGB images, it is still not able to grasp the glass bottles reliably, because they differ so much from the objects it observed during training. However, after fine-tuning with only 1 hour (100 grasp attempts) of experience, we observe that the transparent bottles can be picked up with a success rate of 66\%, 20\% better than the base policy. Figure~\ref{fig:kcam} shows how the robot's view changes for each challenge task. Note the extreme glare and robot reflections visible in images from the Harsh Lighting challenge.

For videos of our experimental results, see the project website.\footnote{For video results, see \url{https://ryanjulian.me/continual-fine-tuning}}

\begin{table*}[t]
    \centering
    \def\arraystretch{0.9}
    \setlength{\tabcolsep}{1em}
    \begin{tabularx}{0.935\linewidth}{lcll*{9}{c}}
        \toprule
        \multicolumn{1}{l}{\multirow{2}[4]{*}{Challenge Task}} & \multicolumn{1}{l}{\multirow{2}[4]{*}{Original Policy}} & \multicolumn{7}{c}{Ours (exploration grasps)} & \multicolumn{2}{c}{Comparisons} \\
        \cmidrule(lr){3-9} \cmidrule(lr){10-11}
        \multicolumn{1}{c}{} & \multicolumn{1}{c}{} & \multicolumn{1}{c}{25} & \multicolumn{1}{c}{50} & \multicolumn{1}{c}{100} & \multicolumn{1}{c}{200} & \multicolumn{1}{c}{400} & \multicolumn{1}{c}{800} & \multicolumn{1}{c}{\textbf{Best ($\Delta$)}} & \multicolumn{1}{c}{Scratch} & \multicolumn{1}{c}{ImageNet} \\
        \midrule
        Checkerboard Backing        & 50\%    & 67\%      & 48\%      & 71\%      & 47\%     & 89\%   & 90\%      & \textbf{90\% (+40)}   & 0\%      & 0\% \\
        Harsh Lighting              & 32\%    & 23\%      & 16\%      & 52\%      & 44\%     & 58\%   & 63\%      & \textbf{63\% (+31)}   & 4\%      & 2\% \\
        Extend Gripper \SI{1}{\cm}  & 75\%    & 93\%      & 67\%      & 80\%      & 51\%     & 90\%   & 69\%      & \textbf{93\% (+18)}   & 0\%      & 14\% \\
        Offset Gripper \SI{10}{\cm} & 43\%    & 73\%      & 50\%      & 60\%      & 56\%     & 91\%   & 98\%      & \textbf{98\% (+55)}   & 37\%     & 47\% \\
        Transparent Bottles         & 49\%    & 46\%      & 43\%      & 65\%      & 65\%     & 58\%   & 66\%      & \textbf{66\% (+17)}   & 27\%     & 20\% \\
        Baseline Grasping Task      & 86\%    & 98\%      & 81\%      & 84\%      & 78\%     & 93\%   & 89\%      & \textbf{98\% (+12)}   & 0\%      & 12\% \\ 
        \bottomrule 
    \end{tabularx}
    \caption{
    Summary of grasping success rates ($N \geq 50$) for the experiments by challenge task, fine-tuning method, and number of exploration grasps. The experiments ``Scratch'' and ``ResNet 50 + ImageNet'' both use 800 exploration grasps and the same update process as the other experiments. ``Scratch'' starts the grasping network with randomly-initialized parameters. ``ResNet 50 + ImageNet'' refers to training a grasping network with an equivalent architecture to the other experiments, but with its convolutional layers replaced with a ResNet 50 architecture and pre-loaded with ImageNet features; the non-CNN parts of the network (MLPs for the action inputs and the Q-value output) are randomly-initialized.}
    \label{tab:final_results}
\end{table*}

\paragraph{Collect datasets}
First, we collect a dataset of 800 grasp attempts for each of our 5 challenge tasks (see Table~\ref{tab:modifications}) plus the base grasping task. We then partitioned each dataset into 6 tiers of difficulty by number of exploration grasps (25, 50, 100, 200, 400, and 800 grasp attempts), yielding 36 individual datasets.

\paragraph{Train fine-tuned policies}
We train a fine-tuned policy for each of these 36 datasets using the procedure described above. We execute the fine-tuning algorithm for 500,000 gradient steps (see Sec.~\ref{sec:analysis} for more information on how we chose this number) and use a learning rate of $10^{-4}$, which is 25\% of learning rate used for pre-training. This yields 36 fine-tuned policies, each trained with a different combination of target task and target dataset size. This set of 36 policies includes 6 policies fine-tuned on data from the base grasping task, for validation.

\paragraph{Train comparisons}
To provide points of comparison, we train two additional policies for each challenge task and the base grasping task, yielding 12 additional policies.

The first comparison (``Scratch'') is a policy trained using the aforementioned fine-tuning procedure and an 800-grasp data set, but using a randomly-initialized Q-function rather than the Q-function obtained from pre-training. The purpose of this comparison is to help us assess the contribution of the pre-trained parameters to the fine-tuning process' performance.

The second comparison (``ImageNet'') is also trained using an identical fine-tuning procedure and the 800-grasp dataset, but uses a modified Q-function architecture in which we replace the convolutional trunk of the network with that of the popular ResNet50 architecture~\cite{he2016deep}, initialized with the weights obtained by training the network to classify images from the ImageNet dataset~\cite{deng2009imagenet}. Refer to to Fig.~\ref{fig:cosines} for a diagram of the unmodified architecture. We initialize the remaining fully-connected layers with random parameters, and concatenate the action input features at the end of the CNN (rather than the adding them in middle of the CNN, as in the original architecture). Note that in this comparison, the fine-tuning process still updates all parameters, including those of the ResNet50 sub-network. The purpose of this comparison is to provide a comparison to a strong alternative to end-to-end RL for obtaining pre-training parameters.

\paragraph{Evaluate performance}
Finally, we evaluate all 48 policies on their target task by deploying them to the robot and executing 50 or more grasp attempts to calculate the policy's final performance. To reduce the variance of our evaluation statistics, we shuffle the contents of the bin between each trial by executing a randomly-generated sequence of sweeping movements with the end-effector.

The full experiment required more than 15,000 grasp attempts and 14 days of real robot time, and was conducted over approximately one month.

We present a full summary of our results in Table~\ref{tab:final_results}. Across the board, we observe substantial benefits arising from fine-tuning, suggesting that the robot can indeed adapt to drastically new condition with a modest amount of data: our most data-intensive experiment uses just 0.2\% of the data used train the base grasping policy to similar performance. Our method consistently outperforms both the ``ImageNet'' and ``Scratch'' comparison methods. We provide more detailed analysis of this experiment in the next section.

The experiments are very challenging. For example, the ``Transparent Bottles'' task presents a major challenge to most grasping systems: the transparent bottles generally confuse depth-based sensors and, especially in cluttered bins, require the robot to singulate individual items and position the gripper in the right orientation for grasping. Although our base policy uses only RGB images, it is still not able to grasp the transparent bottles reliably, because they differ so much from the objects it observed during training. However, after fine-tuning with only 1 hour (100 grasp attempts) of experience, we observe that the transparent bottles can be picked up with a success rate of 66\%, 20\% better than the base policy. Similarly, the ``Checkerboard Backing'' challenge task asks the robot to differentiate edges associated with real objects from edges on an adversarial checkerboard pattern. It never needed this capability to succeed during pre-training, where the background is always featureless and grey, and all edges can be assumed to be associated with a graspable object. After 1 hour (100 grasp attempts) of experience, using our method the robot can grasp objects on the checkerboard background with a 71\% success rate, 21\% better than the base policy, and this success rate reaches 90\% after 8 hours of experience (800 grasp attempts).

\section{Evaluating Offline Fine-Tuning for Continual Learning}
    \begin{figure*}[t]
        \centering
        \noindent
        \includegraphics[width=\textwidth]{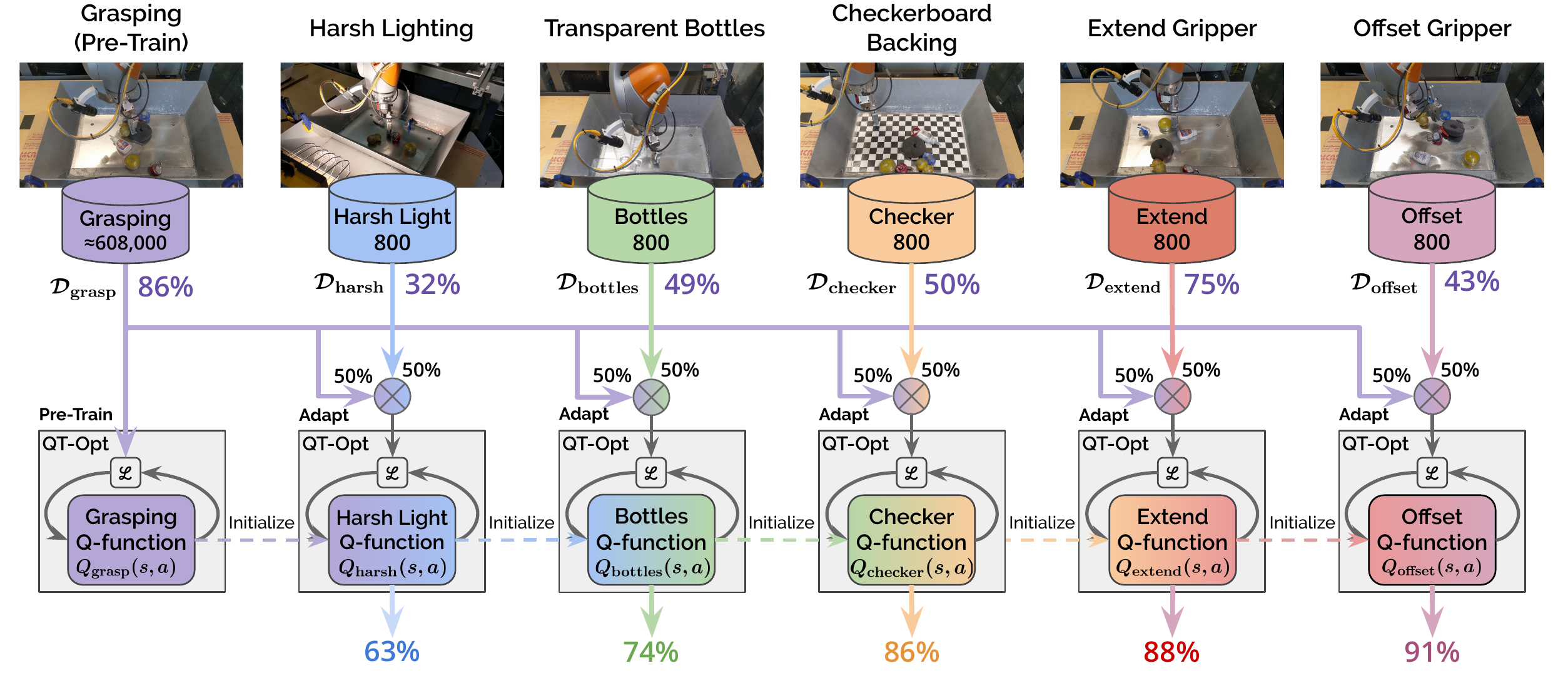}
          \caption{
           Flow chart of the continual learning experiment, in which we fine-tune on a sequence of conditions. Every transition to a new scenario happens after 800 grasps.
          \label{fig:continual_learning}
          }
    \end{figure*}

Now that we have defined and evaluated a simple method for offline fine-tuning, we evaluate its suitability for use in continual learning, which could allow us to achieve the goal of an robot which adapts to ever-changing environments and tasks. To do so, we define a simple continual learning challenge as follows (Fig.~\ref{fig:continual_learning}).

As in the fine-tuning experiments, we begin with a base policy pre-trained for general object grasping. Likewise, we also use our fine-tuning method to adapt the base policy to a target task, in this case ``Harsh Lighting.'' Not content to stop there, we use \textit{this} adapted policy---\textit{not} the base policy---as the initialization for another iteration of our fine-tuning algorithm, this time targeting ``Transparent Bottles.'' We repeat this process until we have run out of new tasks, ending at the task ``Offset Gripper 10cm,'' at which point we evaluate the policy on the last task.

We perform this experiment using 800 exploration-grasp datasets for each Challenge Task from our ablation study of online fine-tuning with real robots. We summarize the results in Table~\ref{tab:cl_results}. Note that because it is the first step of the continual learning experiment, the policy for ``Harsh Lighting'' is identical to that of the 800-grasp variant of the single-step experiment.

\begin{table}[t]
    \centering
    \setlength{\tabcolsep}{1em}
    \begin{tabularx}{1.0\linewidth}{lccc*{10}{c}}
        \toprule
        \multicolumn{1}{c}{\multirow{2}{*}{Challenge Task}} & \multicolumn{1}{c}{\multirow{2}{*}{Continual Learning}} & \multicolumn{2}{c}{$\Delta$} \\
        \cmidrule(lr){3-4}
        \multicolumn{2}{c}{} & \multicolumn{1}{c}{Base} & \multicolumn{1}{c}{Single} \\
        \midrule    
        Harsh Lighting              & $63\%$  & $+32\%$     & $\text{-}$  \\
        Transparent Bottles         & $74\%$  & $+25\%$     & $+8\%$    \\
        Checkerboard Backing        & $86\%$  & $+36\%$     & $-4\%$    \\
        Extend Gripper \SI{1}{cm}   & $88\%$  & $+12\%$     & $-5\%$    \\
        Offset Gripper \SI{10}{cm}  & $91\%$  & $+44\%$     & $-7\%$    \\
        \bottomrule  
    \end{tabularx}
    \caption{
    Summary of grasping success rates ($N \geq 50$) for the continual learning experiment by challenge task, and comparison to single-step fine-tuning. ``Base'' refers to the baseline grasping policy before fine-tuning, and ``Single'' refers to the best performance from the single-step fine-tuning experiment in Table~\ref{tab:final_results}. Note that because it is the first step of the continual learning experiment, the policy for ``Harsh Lighting'' is identical to that of the 800-grasp variant of the single-step experiment.
    }
    \label{tab:cl_results}
\end{table}

Recall that our goal for this experiment is to determine whether continual fine-tuning incurs a significant performance penalty compared to the single-step variant, because we are interested in using this method as a building block for continual learning algorithms. We find that continual fine-tuning does not impose a drastic performance penalty compared to single-step fine-tuning. The continual fine-tuning policies for the ``Checkerboard Backing,'' ``Extend Gripper \SI{1}{cm},'' and ``Offset Gripper \SI{10}{cm},'' challenges succeeded in grasping between 4\% and 7\% less often than their single-step fine-tuning counterparts, whereas the policy for the challenging ``Transparent Bottles'' case actually succeeded 8\% more often. 
These small deltas are within the margin-of-error of our evaluation procedure, so we conclude that the effect of continual fine-tuning on the performance compared to single-step fine-tuning is very small. This experiment demonstrates that our method can perform continual adaptation, and may serve as the basis for a continual end-to-end robot learning method.

\section{Empirical Analysis}
\label{sec:analysis}
In this section, we aim to further investigate the efficiency, performance, and characteristics of our large-scale real-world adaptation experiments. 

\subsection{Performance and sample efficiency of our method}
\begin{figure}
  \noindent
  \centering
  \includegraphics[width=\columnwidth]{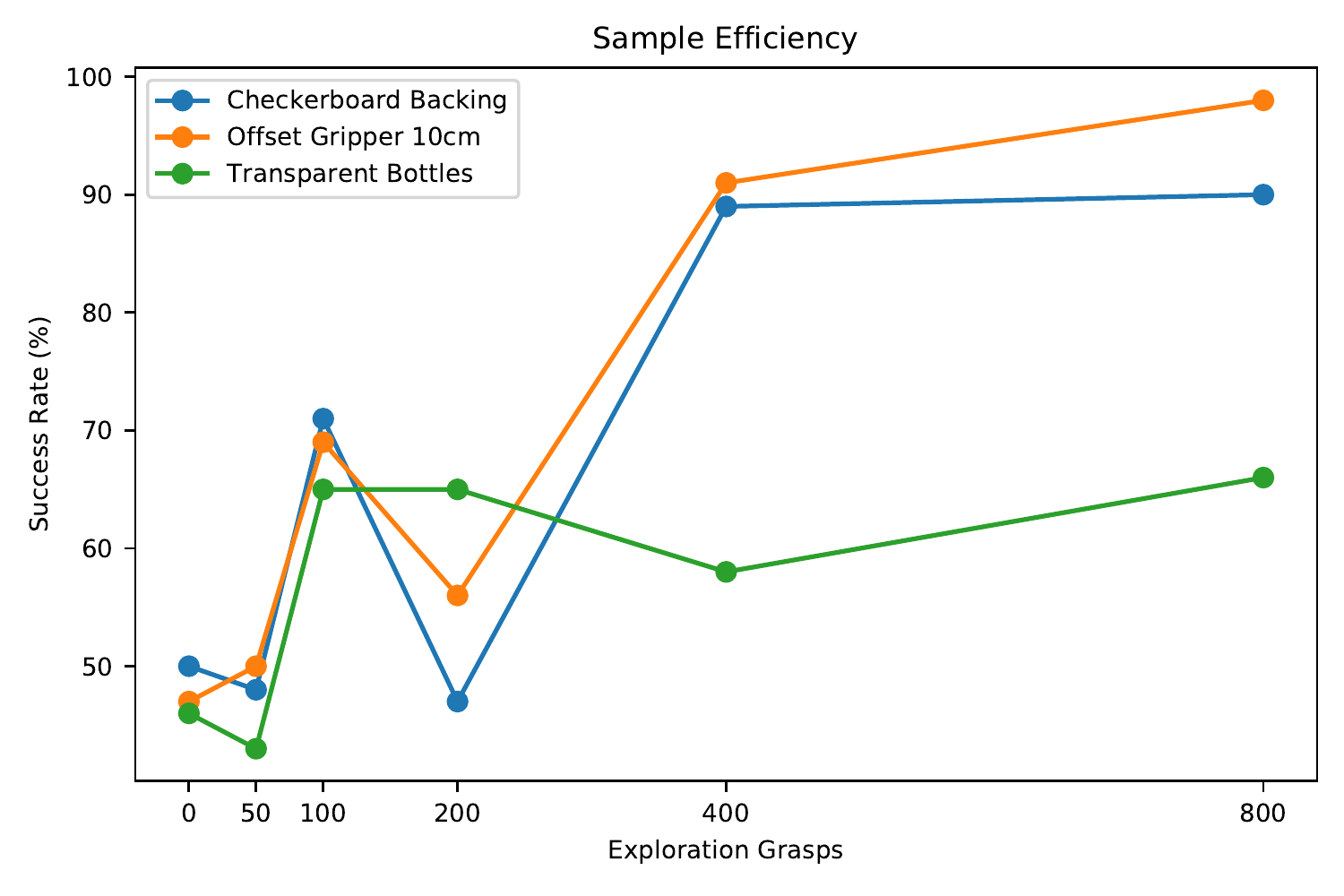}
  \caption{
  Sample efficiency of our fine-tuning method on selected real-robot challenge tasks.}
  \label{fig:sample_efficiency}
\end{figure}
Figure~\ref{fig:sample_efficiency} shows the success rates for our method from Table~\ref{tab:final_results} against the amount of data used to achieve that success rate for selected tasks. The data indicates that our simple offline fine-tuning method can adapt policies to many new tasks with performance at or even above the state-of-the-art base policy, using modest amounts of data. For instance, ``Extend Gripper 1cm'' and ``Offset Gripper 10cm'' both needed only 25 exploration grasps to achieve substantial gains in performance (+18\% and +30\%, respectively). All policies attain substantial performance gains over the base policy by the time they are exposed to 800 exploration grasps, which is less than 0.2\% of the data necessary to train an equivalently-performing policy on the base task.

While the general trend is that more exploration data leads to higher performance, this relationship is not linear. 
All methods experience a substantial improvement in performance after 100 or fewer exploration grasps. 
However, we observe that these performance improvements in the very low-data regime (\eg $\leq 200$ grasp attempts) are also unstable.

\subsection{The downside of offline fine-tuning: deciding when to stop}
\begin{figure}
  \noindent
  \centering
  \includegraphics[width=\columnwidth]{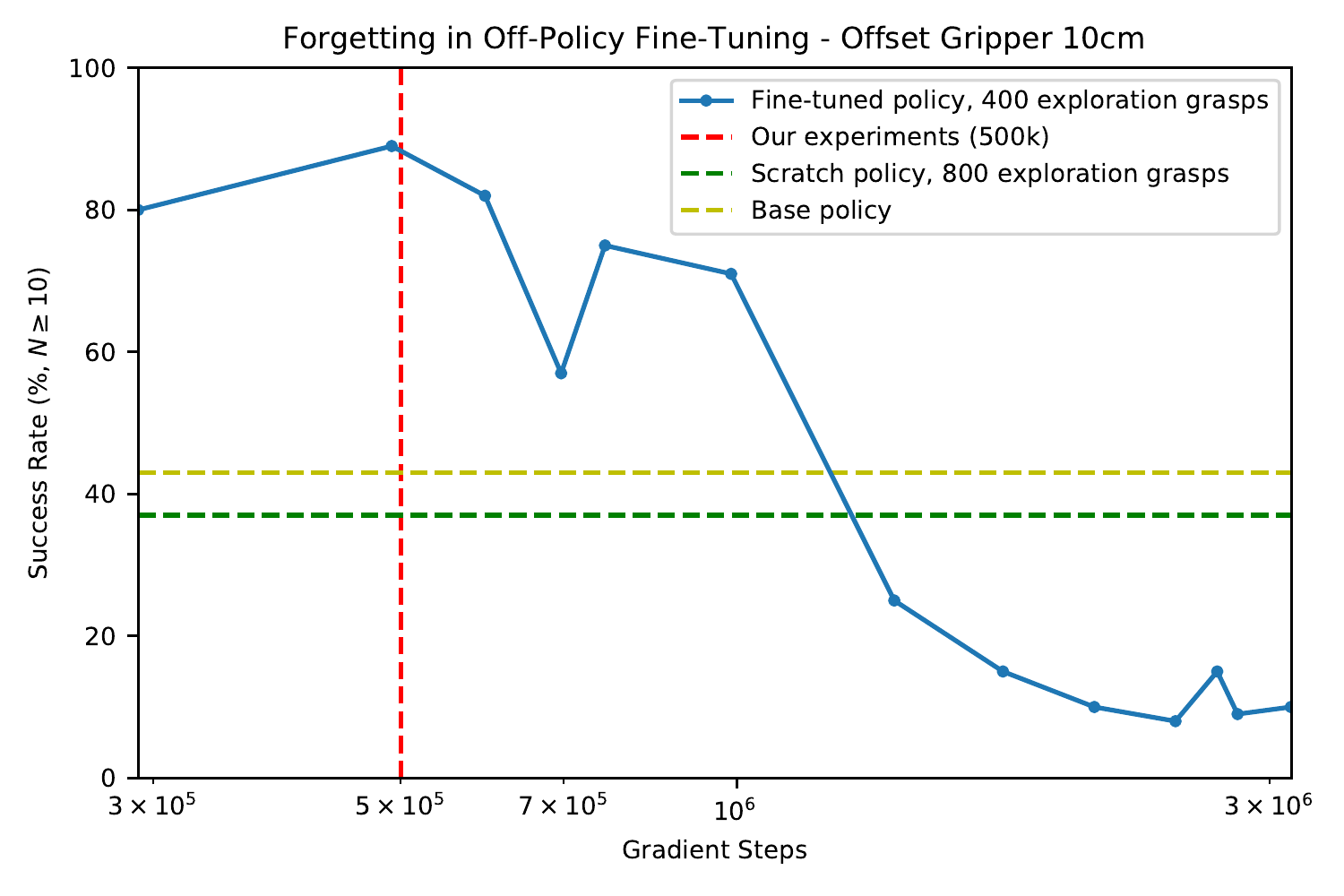}
  \caption{
  Evaluation performance of a single offline fine-tuning experiment at different numbers of gradient steps (optimization epochs). The blue curve is real robot performance on the target task (Offset Gripper 10cm) when trained using 400 exploration grasps. The green dotted line is the performance of training the same policy from scratch (random initialization) using 800 exploration grasps, and the yellow dotted line the performance of the base policy. The red dotted line portrays the number of gradient steps we choose to use for our large-scale fine-tuning study.}
  \label{fig:forgetting}
\end{figure}

Our results indicate that offline fine-tuning can train robotic policies to substantial performance improvements with modest amounts of data, and that offline methods are not limited by the need to preserve an always-sufficient exploration policy as with online methods. However, we identify one significant drawback to the method compared to online fine-tuning.

A pure offline fine-tuning method has no built-in evaluation step which would inform us when the robot's performance on the target task has stopped improving, and therefore when we should stop fine-tuning with a fixed set of target task data. This is a subset of the off-policy evaluation problem~\cite{irpan2019off}. Knowing when the policy stops improving is important, because fine-tuning exists in a low-data regime, and repeatedly updating a neural network model with small amounts of data leads to overfitting onto that data. Not only does this degrade the performance on the target task, but also the ability of the network to adapt to new tasks later (\ie for continual learning).

We can see this phenomenon in Figure~\ref{fig:forgetting} showing a real robot's performance on the ``Offset Gripper 10cm'' target task at different numbers of steps into an offline fine-tuning process that uses 400 exploration grasps. Performance quickly rises until around 500,000 gradient steps. Past this point, it precipitously drops and never recovers, dropping below even the initial performance of the base policy from which it was trained, as the initialization is being overwritten by overfitting to the target samples. The point at which overfitting begins is a function of the initialized model, target dataset, learning algorithm, and many other factors, and is not necessarily stable or easily predictable.

For the purposes of our large-scale fine-tuning study, we use this experiment and several others to determine that 500,000 gradient steps was an acceptable choice for the real-world experiments, but the variance in the results in Table~\ref{tab:final_results} and Figure~\ref{fig:sample_efficiency} shows that this choice was not necessarily optimal for all of our tasks and datasets. 
We believe one practical a solution to this problem of a continual learning robot is to use a mix of offline fine-tuning and online evaluation. The point, at which performance stops improving represents when the training process has exhausted the fine-tuning dataset of new information, and the robot must return to exploring online to continue improving.

\subsection{Comparing initializing with RL to initializing with supervised learning}
\begin{figure*}[t]
  \centering
  \noindent
  \includegraphics[width=\textwidth]{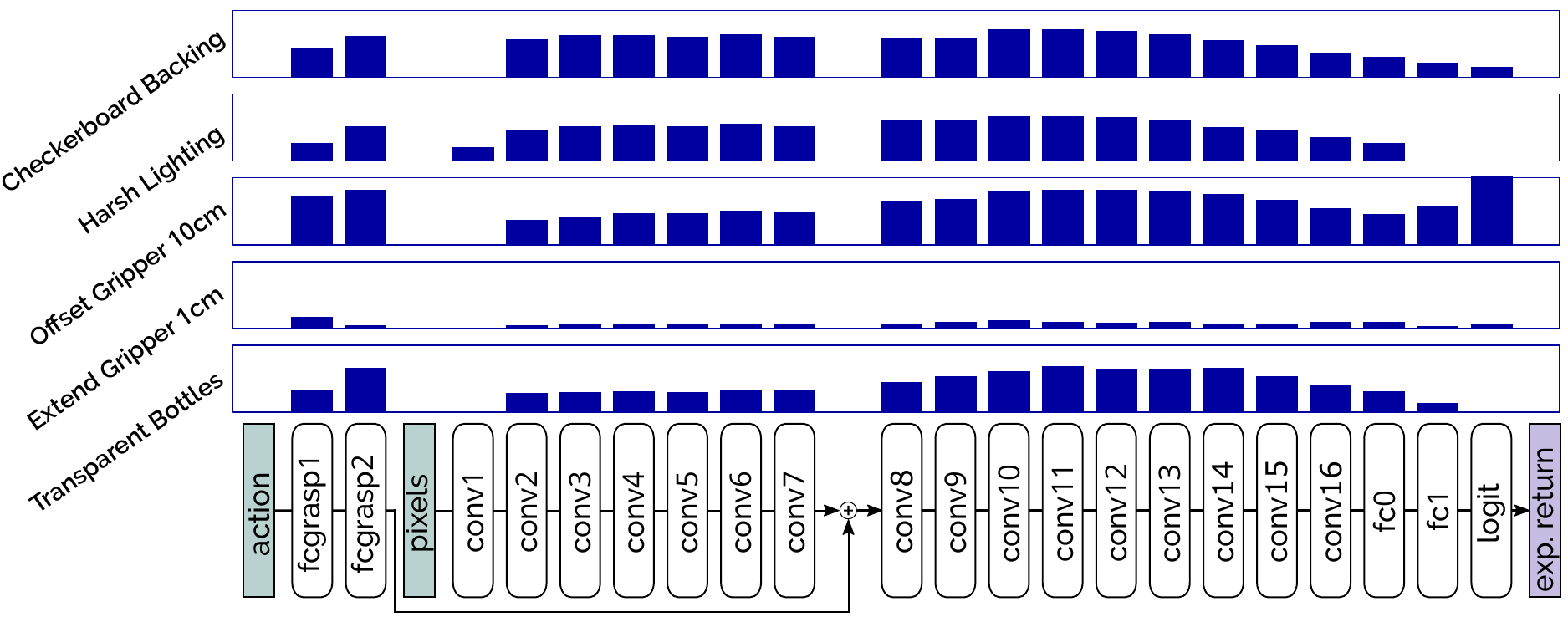}
  \caption{
  Analysis of parameter changes induced by different fine-tuning target tasks. This plot portrays the cosine distance between the parameters of the pre-trained and fine-tuned networks for our 5 fine-tuning target tasks. The bar heights are normalized by the magnitude of parameter changes induced in the Q-function network by fine-tuning the baseline grasping task.}
  \label{fig:cosines}
\end{figure*}

In order to answer the question whether RL is better suited for creating a continually-learning robotic agent than supervised learning, we compare our results to an ImageNet-pretrained baseline. 
The ImageNet baseline uses a similar grasping network where its convolutional layers are replaced with ResNet 50 architecture and pre-loaded with ImageNet features. 
Since the part of the network that process robot's state and action inputs cannot be initialized using supervised learning, we initialize them randomly.
As shown in Table~\ref{tab:final_results}, the best performing ImageNet-based agent achieves the success rate of 47\% on ``Offset Gripper 10cm,'' which corresponds to 4\% improvement over the base policy performance. 
This result seems to confirm our hypothesis that our RL-based pre-training is crucial for good subsequent fine-tuning. Note that we first attempted to fine-tune these ImageNet-based policies while holding the ImageNet feature layers constant, but this procedure failed to achieve any non-zero success rate. This suggests that, unlike adapting computer vision networks to new visual tasks, adapting end-to-end robot learning to new sensorimotor tasks may require changing the features used to represent the problem, and not just the post-processing of said features.

Figure~\ref{fig:cosines} highlights some of the changes that happen during the RL-based fine-tuning in greater detail. 
It demonstrates the normalized distance in parameter space of a fine-tuned policy for each of our challenge tasks from its base policy.
While it is unsurprising that primarily-visual challenges such as ``Checkerboard Backing'' and ``Harsh Lighting'' induce large changes in the parameters of the convolutional parts of the network, we observe that even `Offset Gripper 10cm,'' a purely-morphological change to the robot, induces substantial changes to the network's image-processing parameters (e.g. layers conv2-conv7). 
We attribute this to the successful agent's need for hand-eye coordination to complete the task: offsetting the gripper not only changes robot morphology, it changes the location of the robot in its own visual field drastically. 
In order to perform effective visual servoing with a new morphology, both the image and action-processing parts of the network must be updated.

\section{Conclusion and Future Work}
\label{sec:conclusion}
For robots to be able to operate in unconstrained environments, they must be able to continuously adapt to new situations. We empirically studied this challenge by evaluating a state-of-the-art vision-based robotic grasping system, and testing its robustness to a range of new conditions such as varying backgrounds, lighting conditions, the shape and appearance of objects, and robot morphologies. We found that these new conditions degraded performance of the trained grasping system substantially. Motivated by this initial study, we explored how to adapt vision-based robotic manipulation policies by fine-tuning with off-policy reinforcement learning.

Our large-scale study shows that combining off-policy RL with a very simple fine-tuning procedure is an effective adaptation method, and this method is capable of achieving remarkable improvements in robot performance on new tasks with very little new data. Furthermore, our continual learning experiment shows that using this simple method in a continual setting imposes very little performance penalty compared to the single-step setting. This suggests that the combination of off-policy RL and fine-tuning can serve as a building block for future continual learning methods.

Our results comparing supervised-learning-based initialization to those acquired with our RL-fine-tuning approach highlight a familiar truism about robotics: that robotic agents must do more than perceive the world, they must also act in it. The ability to learn the combination of these two capabilities is what makes RL well-suited for creating continually-learning robots.

While our work demonstrated promising results on a real-world robotic grasping system under a wide range of scenarios, both perceptual and physical, further work is needed to understand how such adaptation performs on a broader range of robotic manipulation tasks. In the future, we would also like to focus on using off-policy metrics such as~\cite{irpan2019off} for the purposes of early stopping, which would allow us to continuously monitor progress of the online fine-tuning process without costly real-robot evaluations. We would also like to further assess our method's suitability for continual adaptation, by assessing its performance on longer continual learning sequences, and measuring the how continual fine-tuning updates for new tasks affects the performance of previously-seen tasks.

\section*{Acknowledgments}
The authors thank Noah Brown and Ivonne Fajardo for their superb and unyielding support with real robot experiments. We also thank Alex Irpan and Eric Jang for their help with robot learning software, Yevgen Chebotar for his advice on early revisions of this work and always-insightful discussions, Dmitry Kalashnikov and Jake Varley for their help with QT-Opt, and K.R. Zentner for her help with editing and artwork for this paper.

\bibliographystyle{plainnat}
\bibliography{references}

\appendix

We propose a conceptual framework for fine-tuning algorithms, and use simulation experiments to assess the suitability of some algorithm variations for end-to-end robot learning.

``Fine-tuning'' refers to a family of transfer learning techniques, in which we seek to acquire a neural network for one task (which we will refer to as the ``target'' task) by making use of some or all of a network trained on a related task (the ``base'' task). This is a is a very common technique for quickly acquiring new tasks in computer vision~\cite{donahue2014decaf,huh2016makes,kornblith2019better} and natural language processing~\cite{howard2018universal}. As collecting new robot experience data is expensive, our goal is to use as little target task data as possible.
In this section, we first describe the general algorithmic sketch for fine-tuning, then enumerate some of the most common fine-tuning techniques. In Sections~\ref{sec:experiments} and~\ref{sec:analysis}, we evaluate the suitability of these techniques for end-to-end robot learning.

\subsection{Fine-Tuning: Conceptual Framework}
\label{sec:fine_tuning_framework}
We can organize fine-tuning for end-to-end reinforcement learning into four essential steps (Fig.~\ref{fig:flowchart}). Different fine-tuning techniques change the details of one of these steps.
\begin{enumerate}
    \item \textbf{Pre-training}: \textit{Pre-train a policy to perform some base task, which is related to our target task.}
    In the experiments in this work, the base task is always indiscriminate object grasping. In computer vision and NLP, this step can often by skipped by making use of one of many pre-trained and publicly-available state-of-the-art vision and language models. We hope for a future in which this is possible in robotics.
    \item \textbf{Exploration}: \textit{Explore in the new target task, to collect data for adaptation.}
    In principle, in off-policy reinforcement learning any policy may be used for exploration. In our study, and what we believe to be most representative of a real-world continual learning scenario, we always use the pre-trained policy for exploration.
    \item \textbf{Initialization}: \textit{Initialize the policy for the target task using some or all of the weights from the pre-trained policy.}
    The standard implementation of this step is to start with the entire pre-trained network. Some techniques may choose to use only a subset of the pre-trained network (\eg truncating the last few layers of a CNN).
    \item \textbf{Adaptation}: \textit{Use the exploration data update the initialized policy to perform the new task.}
    The standard version of this step continues updating the entire initialized policy with the same algorithm and hyperparameters as was used for the pre-training process, but with the target task data. There are many variations on this step, including which parts of the network to update, at what learning rate, with what data, with which optimization algorithm, whether to add additional network layers, etc.
    \item \textbf{Evaluation}: \textit{Assess performance of the fine-tuned network on the new task.}
    If this step only happens once, we refer to such a technique as ``offline fine-tuning,'' because the adaptation step never uses data from an updated policy. If this step happens repeatedly (\eg exploration and evaluation are one-and-the-same), and its result is used for further adaptation to the same target task, we refer to a technique as ``online fine-tuning.'' We explore both variations in our experiments.
\end{enumerate}

Using this fine-tuning framework, we consider several variations of fine-tuning, and assess their suitability for end-to-end robotic RL. Notably, we neglect an analysis of pre-training techniques for fine-tuning reinforcement learning (i.e. (1)), which has a large and rapidly-growing body of research in the meta- and multi-task RL communities (see Sec.~\ref{sec:related_work}). Instead, we focus on initialization (2) and adaptation (3). All of our experiments use end-to-end off-policy reinforcement learning of an indiscriminate object grasping task for their pre-training step. Refer to Section~\ref{sec:pretraining} for details on our pre-training process.

\subsection{Experiments in simulation}
We use simulation experiments to evaluate the suitability of some fine-tuning variations, along the axes we defined in Section~\ref{sec:fine_tuning_framework}.

\subsubsection{Adding a new head and other selective initialization techniques}
\begin{figure}
  \noindent
  \centering
  \includegraphics[width=\columnwidth]{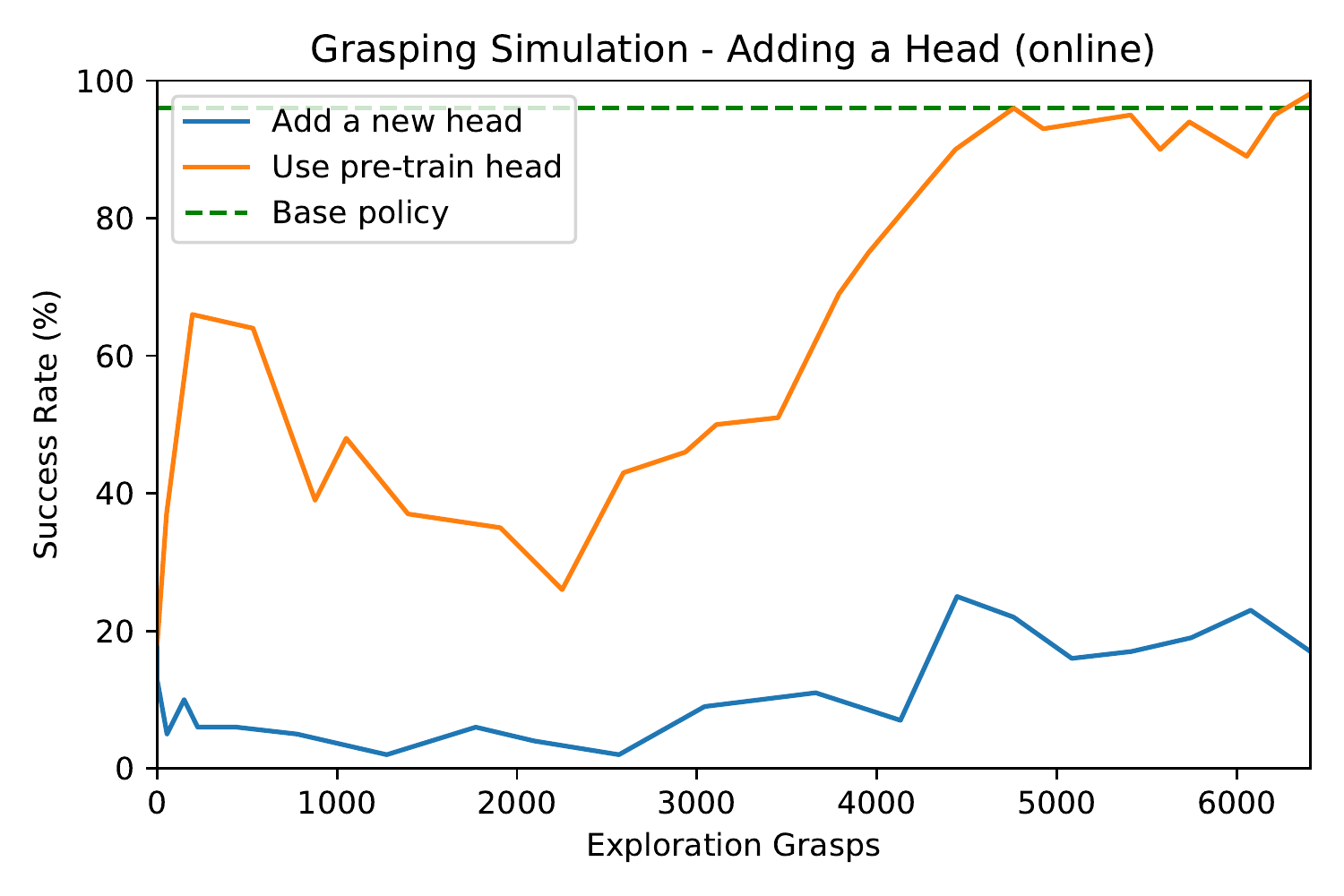}
  \caption{
  Comparison of fine-tuning performance for a policy which uses all base parameters, and a policy which initializes the head parameters from scratch. Re-initializing parameters has a negative effect on sample efficiency for fine-tuning.
  }
  \label{fig:sim_add_head}
\end{figure}

Selective-initialization techniques start the fine-tuning process with a policy which has some of its parameters initialized to random, \eg a popular variant is to ``add a head'' to a pre-trained neural network by omitting its last few layer(s) from initialization, so that the new head can be trained to perform on the target task. 

Figure~\ref{fig:sim_add_head} portrays a study of partial initialization for online fine-tuning using a simulated grasping experiment. In this experiment, the base task is ``grasp opaque blocks'' and the target task ``grasp semi-transparent blocks,'' and the base policy performance is 98\% when trained from scratch on 43,000 grasp attempts. Both fine-tuned policies begin with low performance, around 15\%. After 5000 exploration grasps (12\% of the data used for the base policy), the performance of the full initialization policy has reached the base policy performance, while the policy with a new head has barely reached 30\%. This gap shows that the combination of off-policy RL and selective initialization is unsuitable for sample-efficient fine-tuning.

Our experiments immediately make apparent the downsides of selective initialization for fine-tuning. In particular, online fine-tuning requires to maintain a policy that can competently explore the target task at all times, any method which compromises the performance of such a policy--even temporarily--has a high risk of failing as a sample-efficient fine-tuning technique. The resulting performance gap, once created, is hard to recover from.
As a consequence, we find in simulation experiments that online fine-tuning with selective re-initialization takes a significant fraction of the pre-training samples to converge to baseline performance, making this family of fine-tuning methods sample inefficient.

\subsection{Training with a mix of data from the base and target tasks}
\begin{figure}
  \noindent
  \centering
  \includegraphics[width=\columnwidth]{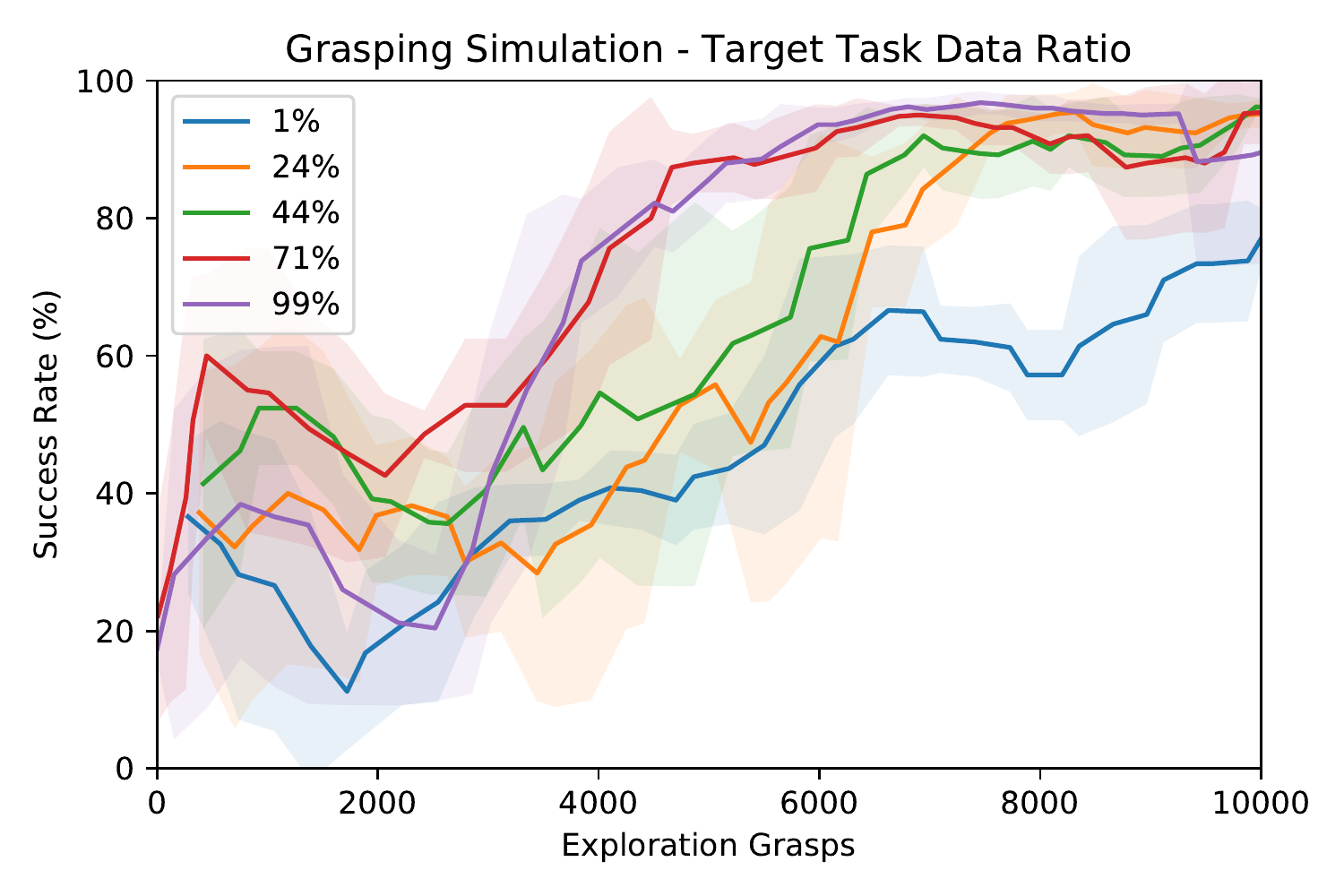}
  \caption{
  Performance curve for an online fine-tuning simulation experiment. The base policy is pre-trained to grasp opaque colored blocks, and the target task is to grasp semi-transparent blocks. Each curve represents a different fraction of target task data, and the remaining data is sampled from the base task. In simulation, the amount of target task data has a straightforward relationship with sample efficiency.}
  \label{fig:sim_ratios}
\end{figure}

We experiment with mixing data from the pre-training task into the fine-tuning process (Fig.~\ref{fig:sim_ratios}), and find that in simulation this has a predictable relationship with sample efficiency: higher shares of target task data allow the fine-tuning policy achieve higher performance faster.

Our goal is to design a fine-tuning algorithm for real robots which might be used for continual learning, and our conclusion from this brief study is that online fine-tuning is a poor fit for for this goal. The experiments with selective re-initialization in particular highlights the challenge of online fine-tuning: it only allows us to use algorithms which preserve the exploration ability of the policy at all times. We also believe that offline fine-tuning is more practical than online fine-tuning, due to the inherent complexity of placing a robot in the loop of a reinforcement learning algorithm. If used as part of a continual learning method, an offline method would also allow a robot to collect data on a new task piecemeal, and only attempt to adapt to that new task when it has collected enough data to be successful.

\end{document}